\documentclass[a4paper]{article}

\usepackage[utf8]{inputenc}
\usepackage{erk}
\usepackage{times}
\usepackage{graphicx}
\usepackage{multirow} 
\usepackage[colorlinks=true, linkcolor=red, citecolor=green]{hyperref}
\usepackage[top=22.5mm, bottom=22.5mm, left=22.5mm, right=22.5mm]{geometry}

\usepackage[slovene,english]{babel}

\def\footnotemark{}%  to avoid footnote on cover page

\begin{document}
%make title
\title{Beyond Detection: Visual Realism Assessment of Deepfakes}

\author{Luka Dragar$^{1}$, Peter Peer$^{1}$,  Vitomir Štruc$^{2}$, Borut Batagelj$^{1}$} % use ^1, ^2 for author(s) from different institutions

\affiliation{$^{1}$Faculty of Computer and Information Science, University of Ljubljana, Večna pot 113, SI-1000 Ljubljana, Slovenia\\$^{2}$University of Ljubljana, Faculty of Electrical Engineering Trzaška cesta 25, 1000 Ljubljana, Slovenia}

\email{E-mail: luka.dragar3@gmail.com}

\maketitle

\begin{abstract}{
Abstract}
In the era of rapid digitalization and artificial intelligence advancements, the development of DeepFake technology has posed significant security and privacy concerns. This paper presents an effective measure to assess the visual realism of DeepFake videos. We utilize an ensemble of two Convolutional Neural Network (CNN) models: Eva and ConvNext. These models have been trained on the DeepFake Game Competition (DFGC) 2022 dataset and aim to predict Mean Opinion Scores (MOS) from DeepFake videos based on features extracted from sequences of frames. Our method secured the third place in the recent DFGC on Visual Realism Assessment held in conjunction with the 2023 International Joint Conference on Biometrics (IJCB 2023). We provide an over\-view of the models, data preprocessing, and training procedures. We also report the performance of our models against the competition's baseline model and discuss the implications of our findings.
\end{abstract}

\section{Introduction}

In the digital era, DeepFakes have emerged as a significant phenomenon in digital media. These AI-generated synthetic videos, focusing on face-swapping, create hyper-realistic counterfeit content that is increasingly challenging to distinguish from authentic media. This development raises major security and privacy concerns, necessitating effective measures for visual realism assessment of DeepFake videos. This paper, in the context of the DeepFake Game Competition on Visual Realism Assessment (DFGC-VRA) 2023, investigates this pressing issue. The ultimate goal of face-swapping is to convince human viewers; therefore, subjective realism assessment plays a critical role. It's vital not only for estimating the potential impact of fake videos on social networks but also for evaluating the performance of face-swapping models during their development.
\begin{figure}[!htb]
    \centering
    \includegraphics[width=0.5\textwidth]{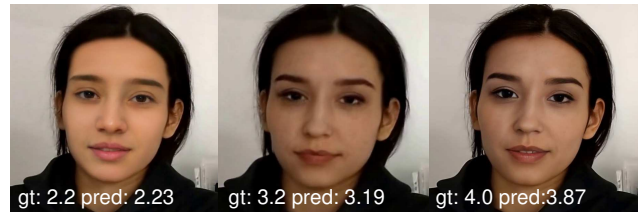}
    \caption{Face-swap videos with different degrees of
realism, annotated with the ground truth MOS (gt) vs.
predicted MOS (pred) by our ensamble model.} 
    \label{slika}
\end{figure}

\section{Related work}
Deepfake detection and realism assessment have been a critical focus of research in recent years due to the rapid development of deepfake technologies. Initial studies primarily targeted deepfake detection, aiming to differentiate authentic videos from AI-manipulated ones. With the advent of deep learning, detection methods have seen significant improvements. However, disparities between human and machine perception of deepfakes, as demonstrated by Korshunov and Marcel~\cite{DBLP:journals/corr/abs-2009-03155}, suggest further research is necessary in this field.

Another key study~\cite{Groh_2021} highlighted the effectiveness of combining human judgement and machine learning models for deepfake detection. The study found that this combination yielded superior performance compared to either humans or the model alone. Interestingly, the study found that the ability to process facial visuals, a specialized cognitive capacity, significantly influenced human deepfake detection performance.

Visual Realism Assessment (VRA) is an extension of the detection problem, focusing not only on the authenticity of videos but also on the quality of the fakes. Recent studies, such as the DeepFake Game Competition on Visual Realism Assessment (DFGC-VRA), have aimed to develop models that predict the Mean Opinion Score (MOS) for deepfake videos, a measure of subjective quality.

Sun et al., in their paper "Visual Realism Assessment for Face-swap Videos,"~\cite{sun2023visual} proposed a benchmark for evaluating the effectiveness of various automatic VRA models. They used the DFGC 2022 dataset for their evaluations and demonstrated the feasibility of creating effective VRA models for assessing face-swap videos. They further emphasized the usefulness of existing deepfake detection features for VRA.

In the DFGC 2022, the winning team, HanChen, used an ensemble of models including ConvNext for their deepfake detection solution~\cite{peng2022dfgc}. This success suggests that models like ConvNext and Vision Transformers can be effectively used not only for deepfake detection but potentially also for Visual Realism Assessment. This underscores the importance of considering visual realism and human cognitive abilities in the development of deepfake detection models and motivates the focus of our current work.

\section{Methods}
Our study leverages an ensemble of two distinctive Convolutional Neural Network (CNN) models: Eva~\cite{fang2022eva} and ConvNext~\cite{liu2022convnet}. Both these models are equipped with tailored regression heads, specifically designed to predict Mean Opinion Scores (MOS) from deepfake videos based on features gathered from seqences of 5 frames. These models have been trained on the DFGC 2022 dataset and represent our contribution 
%the contributions from the University of Ljubljana 
to the recent \textit{DeepFake Game Competition on Visual Realism Assessment}. This competition was held in tandem with the 2023 International Joint Conference on Biometrics (IJCB 2023). Our method managed to secure the third place in this competition.

The code for the implementation of our models, alongside a detailed technical report, is publicly available on our GitHub repository\footnote{https://github.com/TheLukaDragar/UNI-LJ-VRA}.

The ConvNext~\cite{liu2022convnet} model we employ is part of a family of convolutional neural networks (ConvNets). It is an evolution of the ResNet architecture, progressively incorporating elements from hierarchical vision Transformers. This model is initialized with weights from the winner of the previous year's DFGC2022 competition, the DFGC-1st-2022-model~\cite{peng2022dfgc}.

The architecture of the ConvNext model leverages the pre-trained DFGC-1st-2022-model as its backbone. This model, once loaded, is used to extract output features from the video frames. The last fully connected layer of the backbone model is replaced with an identity layer, effectively removing it from the model architecture. A dropout layer is then introduced to manage model complexity.

A defining feature of our ConvNext model is its me\-thod of feature aggregation, specifically designed to handle the mean and standard deviation vectors from video frames. To accommodate these aggregated features, several fully connected layers are added. The final layer in this structure is specifically designed to output the Mean Opinion Score (MOS).

The forward pass of the ConvNext model begins with an input video sequence. A random starting point is chosen within the video, from which a sequence of 5 consecutive frames is selected. Each frame is processed by the backbone model to extract its features.

Following this, the mean and standard deviation of these features are calculated for each frame sequence. The calculation is guided by the principles of average pooling and standard deviation pooling, common techniques in the video quality assessment field. Specifically, the mean \(f_{\mathrm{mean}}\) and standard deviation \(f_{\mathrm{std}}\) of the features are computed as follows:

\begin{equation}
f_{\mathrm{mean}} = \frac{1}{n}\sum_{i=1}^{n} f_{i}
\label{eq:fmean}
\end{equation}

\begin{equation}
f_{\mathrm{std}} = \sqrt{\frac{1}{n - 1}\sum_{i=1}^{n}(f_{i} - f_{\mathrm{mean}})^2}
\label{eq:fstd}
\end{equation}
Here, $n$ represents the total number of frames. The feature vector for the $i$-th frame is denoted as \(f_i\) and \(f_{\mathrm{mean}}\) is the average feature vector.

These computed mean and standard deviation vectors are concatenated to form the video-level features. They are then fed into the fully connected layers of the ConvNext model, culminating in the final output: the Mean Opinion Score (MOS) for the input frame sequence.

%Notably, the weights of the backbone model are not frozen in the ConvNext model. 
It is noteworthy that the weights of the backbone mo\-del are not frozen in the ConvNext mo\-del during training.
This approach allows for the fine-tuning of the entire model, including the backbone and the newly added layers, for the specific task of predicting MOS scores on deepfake videos. The learning objective used is the Root Mean Squared Error (RMSE), which provides a measure of the differences between values predicted by the model and the values actually observed.

In parallel to the ConvNext model, we also utilize the Eva~\cite{fang2022eva} model, a novel approach to visual representation learning that explores the limits of large-scale masked image modeling (MIM) using publicly accessible data. Eva is a vanilla Vision Transformer (ViT) that is pre-trained to reconstruct masked out image-text aligned vision features conditioned on visible image patches. This unique pre-training task allows Eva to scale efficiently, even up to one billion parameters, setting new records across a wide range of downstream visual tasks, such as image recognition, video action recognition, object detection, instance segmentation, and semantic segmentation, without heavy supervised training.

Just as with ConvNext, we use Eva as a feature extractor for predicting Mean Opinion Scores (MOS) on deepfake videos. The architecture is similar to that of ConvNext, with the backbone model replaced by the Eva model. However, unlike ConvNext, the Eva model was not initialized with weights from the previous year's winners of the DFGC2022 competition. 
Instead, we used pretrained weights on ImageNet using the timm library\footnote{https://github.com/rwightman/pytorch-image-models/tree/main/results (timestamp: Jun 8, 2023).}.
%Instead, we used pretrained weights from timm library \footnote{https://github.com/rwightman/pytorch-image-models/tree/main/results (timestamp: Jun 8, 2023).}.

\section{Experiments}
\subsection{Dataset}
The DeepFake Game Competition (DFGC) 2022 dataset, which originated from the second DFGC held in conjunction with the IJCB-2022 conference, is composed of 2799 fake and 1595 real face-swap videos, each approximately 5 seconds in length. The fake videos are generated using a variety of face-swap methods like DeepFaceLab~\cite{perov2021deepfacelab}, SimSwap~\cite{Chen_2020}, and FaceShifter~\cite{li2020faceshifter}, along with post\-pro\-cess\-ing operations.

These videos are divided into three subsets: C1, C2, and C3, each containing different amounts of data. Specifically, C1 includes 240 fake videos from 6 submit-ids, C2 contains 520 fake videos from 13 submit-ids, and C3 houses 640 fake videos from 16 submit-ids. Each submit-id corresponds to a set of 80 swap videos for 20 pairs of facial-IDs, with videos from the same submit-id believed to be created by identical methods or processes.

The dataset also includes annotations by five independent human raters who assessed video realism among other factors, rating it on a scale of 1 (very bad) to 5 (very good) (Figure~\ref{slika}).
\subsection{Preprocessing}

The data preprocessing stage of our work with the DFGC-2022 dataset involved extracting faces from each video frame with the Multi-task Cascaded Convolutional Networks (MTCNN)~\cite{Zhang_2016} model and OpenCV~\cite{opencv_library}. The bounding boxes of the faces were resized and adjusted by a scale factor of 1.3, providing context and improving prediction accuracy~\cite{peng2022dfgc}.

This process generated a new dataset of cropped face images, used as input for subsequent modeling. By performing preprocessing prior to model training, we were able to expedite the training process, saving time and computational resources.

\subsection{Training data}
We implemented a method to process each video by selectively extracting sequences of frames. This involved randomly selecting a starting point within each video and subsequently capturing a sequence of five frames from that point. The frames were then transformed in accordance with the requirements of our modeling process. Each sequence was matched with its corresponding Mean Opinion Score (MOS) label.

Following this, the dataset was divided into three subsets to facilitate the training, validation, and testing of our models. The distribution of the dataset was as follows: out of the total 700 videos, 70\% (490 videos) were allocated for training, 20\% (140 videos) for testing, and the remaining 10\% (70 videos) were used for validation.

\subsection{Experimental setup}
Our training procedure involved the use of a High-Per\-for\-mance Computing (HPC) infrastructure, facilitated by Pytorch Lightning. The employed loss function was Root Mean Square Error (RMSE), and AdamW was chosen as the optimizer with a learning rate of 2e-5. The learning rate scheduler, ReduceLROnPlateau, was inco\-rpora\-ted alo\-ng with early stopping, both of which were monitored via validation loss.

The selected hyperparameters were a batch size of 2, a dropout rate of 0.1, a sequence length of 5, gradient accumulation across 8 batches, and a maximum of 33 epochs as determined by early stopping. Training was conducted on two Tesla V100S-32GB GPUs using a distributed data parallel (ddp) strategy.

Due to the non-deterministic nature of the training process, multiple models were trained with identical parameters. From these runs, the model with the best performance, as determined by validation loss, was selected for the final submission.

Weights \& Biases was utilized for logging and real-time tracking of the training process. At the end of the training, the best model checkpoint based on validation loss was selected and further trained with the same hyperparameters to accommodate the remaining data. Once early stopping was triggered, this final model checkpoint was saved for the final predictions. Access to the details of the best and final runs for both models can be provided via the respective Weights \& Biases links in the tehnical report.

\subsection{Predictions and Model Averaging}
The final model checkpoints of both the ConvNext and Eva models were utilized to make predictions on the test sets for our final submission. These test sets, referred to as Test Set 1 (300 videos), Test Set 2 (280 videos), and Test Set 3 (120 videos), were extracted from different subsets of the DeepFake Game Competition (DFGC) 2022 dataset.

Our dataloader operates in a stochastic manner, selecting sequences of 5 frames from the videos at random. To mitigate the variance introduced by this randomness, we applied an averaging strategy for each test set: we generated predictions 10 times and then computed the average. This procedure yields more robust predictions that accommodate the inherent randomness of our frame sequence selection process.

For the combination of predictions from the two models, we employed a weighted average approach. The final prediction was computed as 0.75 times the ConvNext prediction plus 0.25 times the Eva prediction. This weighting scheme was chosen due to the ConvNext model's superior performance during the training phase.

To assess the consistency of our predictions, we calculated the Root Mean Square Error (RMSE) between pairs of predictions. For the ConvNext model, the average RMSE was found to be 0.16, indicating a reasonable level of consistency in our predictions.

\section{Results}
\begin{figure}[!htb]
    \centering
    \includegraphics[width=0.5\textwidth]{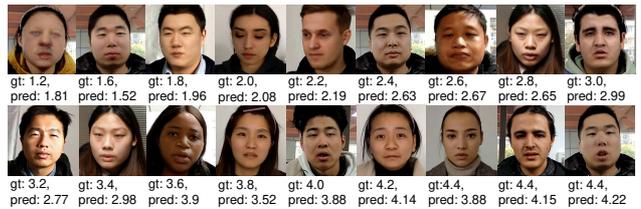}
    \caption{Videos featuring varying levels of realism in face-swapping, annotated according to the Mean Opinion Score (MOS) based on ground truth (gt) and the MOS predicted (pred) by our ensemble model. These videos are arranged in ascending order according to the ground truth MOS.} 
    \label{slika2}
\end{figure}

Our experimental outcomes, are detailed in Tables~\ref{table1} and \ref{table2}, and visually depicted in Figures~\ref{slika} and \ref{slika2}. 
Table~\ref{table1} displays the performance of our two distinct models, Eva and ConvNext, and our ensemble model across three test sets. Each model's performance is evaluated using three metrics: Pearson Linear Correlation Coefficient (PLCC), Spearman Rank-Order Correlation Coefficient (SRCC), both being the competition's official metrics, and Root Mean Square Error (RMSE).

In Table~\ref{table2}, we compare the final scores of our models against the baseline model established by the competition organizers~\cite{sun2023visual}. These final scores, computed by averaging the PLCC and SRCC, provide a comprehensive measure of each model's effectiveness in visual realism assessment of DeepFake videos. As shown in the table, our ensemble model demonstrated a competitive performance and secured a third-place ranking in the competition.

\begin{table}[h]
\caption{Performance Metrics for Each Model on the Test Sets.}
\label{table1}
\smallskip
\begin{center}
\begin{tabular}{ | c | c | c | c | c | }
\hline  
  \textbf{Model} & \textbf{Test Set} & \textbf{PLCC$^{\uparrow}$} & \textbf{SRCC$^{\uparrow}$} & \textbf{RMSE$^{\downarrow}$}\\ 
\hline  
  \multirow{3}{*}{Eva} & 1 & \textbf{0.8305} & \textbf{0.7919} & \textbf{0.4128} \\
                       & 2 & 0.9158 & 0.9119 & 0.3622 \\
                       & 3 & \textbf{0.8726} & \textbf{0.8285} & \textbf{0.4132} \\
\hline
  \multirow{3}{*}{ConvNext} & 1 & 0.7899 & 0.7387 & 0.4545 \\
                            & 2 & 0.9279 & 0.9171 & 0.3492 \\
                            & 3 & 0.8647 & 0.8211 & 0.4303 \\
\hline
  \multirow{3}{*}{Ensemble} & 1 & 0.8091 & 0.7633 & 0.4352 \\
                            & 2 & \textbf{0.9287} & \textbf{0.9197} & \textbf{0.3447} \\
                            & 3 & 0.8746 & 0.8318 & 0.4146 \\
\hline  
\end{tabular}
\end{center}
\end{table}
\begin{table}[h]
\caption{Final Scores for Each Model.} \label{table2}
\smallskip
\begin{center}
\begin{tabular}{ | c | c | }
\hline  
  \textbf{Model} & \textbf{Final Score} \\ 
\hline
  Baseline & 0.5470 \\
\hline  
  Eva & \textbf{0.8585} \\
\hline
  ConvNext & 0.8432 \\
\hline
  Ensemble & 0.8545 \\

\hline  
\end{tabular}
\end{center}
\end{table}

Regarding the data presented in Tables \ref{table1} and \ref{table2}, a noteworthy finding arises in relation to the performance of our Eva model. Despite initially achieving marginally lower results than ConvNext during the training phase, Eva demonstrated superior performance on test sets 1 and 2, ultimately attaining a slightly higher overall score. This indicates that Eva exhibits enhanced generalization capabilities compared to ConvNext.

\section{Conclusion}
In this paper, we investigated the demanding task of evaluating the visual realism of DeepFake videos, by participating in the DFGC-VRA 2023 challenge. Our unique approach involved the use of two distinct convolutional neural network models, ConvNext and Eva. These models were specifically trained to predict Mean Opinion Sco\-res (MOS) of deepfake videos, utilizing the DFGC 2022 dataset. Further enhancements to the models were realized through the integration of strategic pre-processing, feature extraction, and model averaging techniques.

Our experimentation on three distinct test sets demonstrated the promising performance of our models. Most notably, our ensemble model successfully secured a third-place ranking in the DFGC-VRA 2023, highlighting its effectiveness in evaluating visual realism in DeepFake videos.

In summary, this research makes an important contribution to the evaluation of visual realism in deepfake videos, supporting the pursuit of a safer digital media landscape.

%In conclusion, this research contributes significant insights to the visual realism assessment in Deep\-Fake vide\-os, thereby aiding the pursuit of a safer digital media landscape.

\small
\begin{thebibliography}{1}

\bibitem{DBLP:journals/corr/abs-2009-03155}
Pavel Korshunov and S{\'{e}}bastien Marcel,
\textit{Deepfake detection: humans vs. machines},
arXiv preprint arXiv:2009.03155, 2020.

\bibitem{Groh_2021}
Matthew Groh, Ziv Epstein, Chaz Firestone, and Rosalind Picard,
\textit{Deepfake detection by human crowds, machines, and machine-informed crowds},
Proceedings of the National Academy of Sciences, Vol. 119, No. 1, 2021.

\bibitem{sun2023visual}
Xianyun Sun, Beibei Dong, Caiyong Wang, Bo Peng, Jing Dong,
\textit{Visual Realism Assessment for Face-swap Videos},
arXiv preprint arXiv:2302.00918, 2023.

\bibitem{peng2022dfgc}
Bo Peng, Wei Xiang, Yue Jiang, Wei Wang, Jing Dong, Zhenan Sun, Zhen Lei, Siwei Lyu,
\textit{DFGC 2022: The Second DeepFake Game Competition},
arXiv preprint arXiv:2206.15138, 2022.



\bibitem{perov2021deepfacelab}
Ivan Perov, Daiheng Gao, Nikolay Chervoniy, Kunlin Liu, Sugasa Marangonda, Chris Umé, Mr. Dpfks, Carl Shift Facenheim, Luis RP, Jian Jiang, Sheng Zhang, Pingyu Wu, Bo Zhou, Weiming Zhang,
\textit{DeepFaceLab: Integrated, flexible and extensible face-swapping framework},
arXiv preprint arXiv:2005.05535, 2021.

\bibitem{liu2022convnet}
  Zhuang Liu, Hanzi Mao, Chao-Yuan Wu, Christoph Feichtenhofer, Trevor Darrell, Saining Xie,
  \textit{A ConvNet for the 2020s},
  2022.
\bibitem{fang2022eva}
Yuxin Fang, Wen Wang, Binhui Xie, Quan Sun, Ledell Wu, Xinggang Wang, Tiejun Huang, Xinlong Wang, Yue Cao.
\textit{EVA: Exploring the Limits of Masked Visual Representation Learning at Scale},
2022.

\bibitem{Chen_2020}
Renwang Chen, Xuanhong Chen, Bingbing Ni, and Yanhao Ge,
\textit{Simswap: An efficient framework for high fidelity face swapping},
in Proceedings of the 28th ACM International Conference on Multimedia, 2020.

\bibitem{li2020faceshifter}
Lingzhi Li, Jianmin Bao, Hao Yang, Dong Chen, and Fang Wen,
\textit{FaceShifter: Towards High Fidelity And Occlusion Aware Face Swapping},
arXiv preprint arXiv:1912.13457, 2020.

\bibitem{Zhang_2016}
Kaipeng Zhang, Zhanpeng Zhang, Zhifeng Li, and Yu Qiao,
\textit{Joint Face Detection and Alignment Using Multitask Cascaded Convolutional Networks},
\textit{IEEE Signal Processing Letters},
vol. 23, no. 10, pp. 1499--1503, Oct. 2016.


\bibitem{opencv_library}
G. Bradski,
\textit{The OpenCV Library},
\textit{Dr. Dobb's Journal of Software Tools},
2000.



\end{thebibliography}
\end{document}